\documentclass[letterpaper]{article} % DO NOT CHANGE THIS
\usepackage{aaai20}  % DO NOT CHANGE THIS
\usepackage{times}  % DO NOT CHANGE THIS
\usepackage{helvet} % DO NOT CHANGE THIS
\usepackage{courier}  % DO NOT CHANGE THIS
\usepackage[hyphens]{url}  % DO NOT CHANGE THIS
\usepackage{graphicx} % DO NOT CHANGE THIS
\usepackage{graphicx}  % DO NOT CHANGE THIS
\title{Silhouette-Net: 3D Hand Pose Estimation from Silhouettes}
\usepackage{amsmath}
\usepackage{amssymb}
\usepackage{pifont}
\newcommand{\xmark}{\ding{55}}
\newcommand{\cmark}{\ding{51}}

\author{Kuo-Wei Lee,\textsuperscript{1} Shih-Hung Liu,\textsuperscript{1}
Hwann-Tzong Chen,\textsuperscript{1}
Koichi Ito\textsuperscript{2}\\
\textsuperscript{1}{Department of Computer Science, National Tsing Hua University}\\
\textsuperscript{2}{Graduate School of Information Sciences, Tohoku University}}

\begin{document}

\maketitle

\begin{abstract}
3D hand pose estimation has received a lot of attention for its wide range of applications and has made great progress owing to the development of deep learning. Existing approaches mainly consider different input modalities and settings, such as monocular RGB, multi-view RGB, depth, or point cloud, to provide sufficient cues for resolving variations caused by self occlusion and viewpoint change. In contrast, this work aims to address the less-explored idea of using minimal information to estimate 3D hand poses. We present a new architecture that automatically learns a guidance from implicit depth perception and solves the ambiguity of hand pose through end-to-end training. The experimental results show that 3D hand poses can be accurately estimated from solely {\em hand silhouettes} without using depth maps. Extensive evaluations on the {\em 2017 Hands In the Million Challenge} (HIM2017) benchmark dataset further demonstrate that our method achieves comparable or even better performance than recent depth-based approaches and serves as the state-of-the-art of its own kind on estimating 3D hand poses from silhouettes. 
\end{abstract}

%%%%%%%%% BODY TEXT
%===========================================================================
\section{Introduction}
\noindent 
Human hand analysis is an important topic due to its breadth of applications, including AR/VR, object manipulations, driving assistance, and human-computer interaction. In the past decade, studies on vision-based human hand analysis have opened various research directions like 3D hand pose estimation, hand modelling and rendering, gesture recognition, etc. Among these topics, hand pose estimation can be construed as the essence of hand analysis since hand articulation is the core for solving most of the problems. 

% Motivation
Methods on 3D hand pose estimation adopt monocular RGB~\cite{oberweger2015hands}, multi-view RGB~\cite{zhou2016model,ge20173d}, or depth map as the input modality for resolving variations caused by self occlusion~\cite{yuan2018depth} and viewpoint change. Although depth information is supposedly useful for hand pose prediction, it is seldom investigated whether high-resolution or dense depth maps are indeed indispensable. The requirement of high-cost depth sensors for those approaches also constrains the applicability of 3D hand pose estimation. It would be better if such limitations can be relaxed. This motivates us to build a robust and relatively low-cost system for 3D hand pose estimation by reducing the input signals to minimal requirements while retaining comparable accuracy.

\begin{figure}[t]
\centering
\includegraphics[width=0.95\columnwidth]{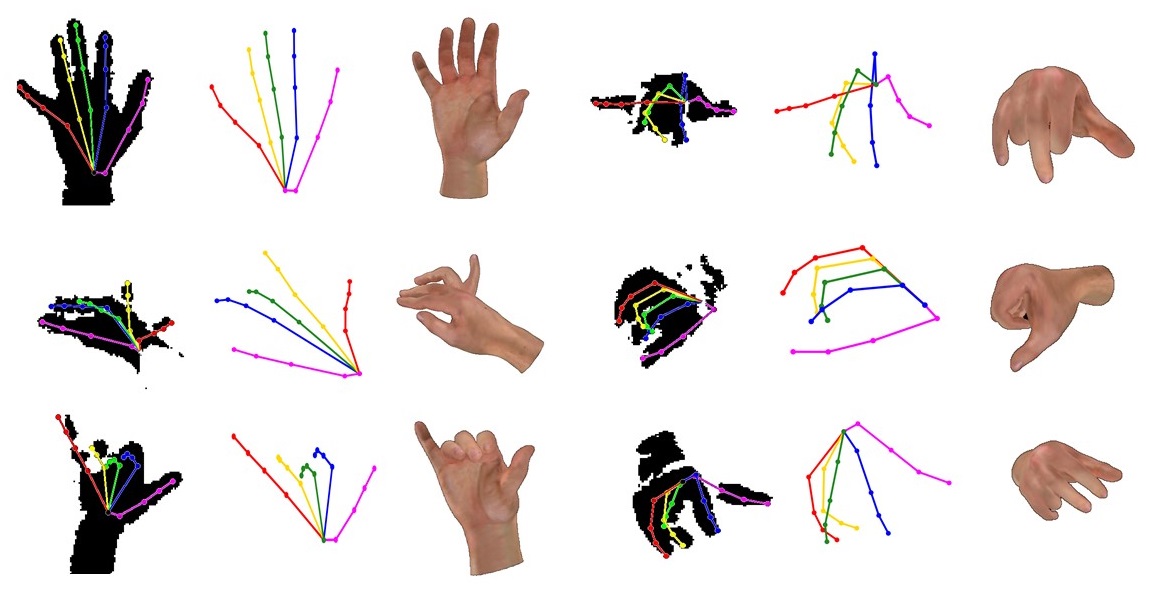} % Reduce the figure size so that it is slightly narrower than the column. Don't use precise values for figure width.This setup will avoid overfull boxes. 
\begin{tabular}{p{7mm}p{9mm}p{13mm}p{7mm}p{9mm}p{11mm}}
Input & Output & Rendered &
Input & Output & Rendered 
\end{tabular}

\caption{The proposed {\em Silhouette-Net} is capable of estimating 3D hand joints from hand silhouettes. We evaluate our model on the dataset of {\em 2017 Hands In the Million Challenge} (HIM2017)~\cite{yuan20172017}, but use simply the binary-valued hand silhouettes as the input. The max-per-joint error achieved by our model is only 7.63mm on the test set. We use the predicted 3D hand poses to render synthetic hand shapes for better visualization. }
\label{fig:teaser}
\end{figure}

%To build a robust system for 3D hand pose estimation, several major challenges must be tackled, for example, unseen hand shapes, extreme viewpoints, and joint occlusions~\cite{yuan2018depth}. 

% What we do in this paper 
We advocate that, for 3D hand pose estimation, accurate prediction can be achieved using limited input from low-cost devices. Our experiments show that the learning of depth perception is comparably more essential than the explicit depth values provided in the depth map. This argument can be supported by the fact that human beings are able to guess the possible hand pose from its silhouette even in an extremely dim environment. Furthermore, the human visual system enables us to perceive relative depth from various visual cues without the need of sensing the actual depth value. These simple facts motivate us to come up with a more challenging task: {\em learning to infer 3D hand joint positions from minimal input information}. We consider the silhouettes obtained by binarizing depth values to a minimal level (i.e., 2D binary maps) as the input for hand pose estimation. At the training phase, the learning model incorporates depth maps as guidance for learning depth perception. However, at the test phase, the system requires nothing but simply the silhouettes of hands to infer the 3D positions of hand joints. We propose a solution to this task and verify the effectiveness of the concept. Some of our 3D hand pose estimation results are shown in Figure~\ref{fig:teaser}.

%__________________________Previous paragraph________________________________%
%It is not surprising that depth-based methods usually perform better than RGB-based approaches since the depth cue should be more explicitly correlated with the 3D positions of joints than the color cue. However, the issue of occlusion still lies in most depth-based approaches. Some recent works apply manually designed constraints to the structure~\cite{oberweger2017deepprior++,Wu18HandPose} and indeed improve the reliability of prediction. Although the limitation of using only the minimal information to estimate hand poses might degrade the prediction accuracy, we design a guidance to overcome this issue. The automatically learned guidance can be seen as a generalized constraint, alleviating the problem of joints occlusion and lack of depth cue for the binary-based approach, which has not been studied yet. 
%____________________________________________________________________________%

% Reviewer2 !!!!! Depth!!!! Underlying  problems, and how we solved.
% Details ( Our work is based on ...
Unsurprisingly, depth-based methods usually perform better than RGB-based approaches in that the depth cue is more explicitly correlated with the 3D positions of joints than the color cue. Silhouette alone holds the relevant structural information, but we still need to address the issues of hand joint occlusion and hand shape variability. To disambiguate the possible poses and predict hand articulations precisely, we design a self-guidance network that learns the implicit depth perception to automatically generate missing spatial features. This self-guidance mechanism alleviates the problems caused by joints occlusion and lack of depth cue. 
%The binary-based approach has not been studied yet. 

% Challenge 
%The major challenge of the proposed approach is avoiding the overfitting problem of using only binary images. Without the depth information, the network could generate some redundant intermediate representations to directly fit the target prediction rather than carefully train the deep network. Moreover, the minimal information of binary image limits the extraction of features to produce heatmaps in the direct regression fashion. Some generative approaches~\cite{wan2017crossing} apply GAN or VAE frameworks to an RGB image and generate a corresponding depth map for the mitigating problems of perspective distortion, occlusion, viewpoints, etc. Other recent works ~\cite{moon2018v2v} cast the hand pose estimation problem into a voxel-to-voxel prediction problem. The existence of 3D point cloud furnish with the sufficient information to address the problem, nevertheless, voxelizing from a binary image is another even harder task. 

%Our work is based on an end-to-end trainable network structure for the inference of 3D hand pose from binary hand images. Specifically, we introduce a latent representation with multi-scale resolution to guide the prediction of hand poses from averting haphazard fitting issue. We refer to the latent representation as the \textit{depth perception} guidance.  

We aim to justify that learning the `concept of depth' (we call it {\em depth perception} in this work) is crucial for improving the accuracy of estimating 3D hand poses from binary-valued silhouettes, given that we do not want to rely on inputs of explicit depth values. Therefore, we propose a novel network architecture, which contains Depth Perceptive Network (DPN) and Residual Prediction Network (RPN). Specifically, DPN guides RPN and yields a combination of detection-based and regression-based approaches. We adopt a structure similar to Feature Pyramid Network (FPN)~\cite{lin2017feature} and U-Net~\cite{ronneberger2015u} to construct DPN, which hierarchically upsamples and downsamples at different scales with the lateral connections in between. With the generated representation and guidance from DPN, our binary-based method can accurately predict 3D hand poses from silhouettes while achieving similar or better performance as depth-based methods.

% Moreover, the designed network architecture incorporate a mechanism of learning depth perception so that it can avoid overfitting by automatically generating a useful latent representation of depth perception and hence can improve the accuracy of heatmap for detecting joint locations.  ( Seems a useless sentence )
To the best of our knowledge, this is the first work to successfully show that estimating 3D hand poses from silhouettes can achieve comparable performance as depth-based methods in terms of mean error and max-per-joint error. Furthermore, inspired by Ge et al.~\cite{ge2016robust}, we design a latent representation to characterize depth perception from multiple views.
%The latent representation contains $C_\mathrm{dim}$ channels, where $C_\mathrm{dim}$ equals the number of hand joints times the number of views. 
The main contributions of this work can be summarized as follows:
\begin{enumerate}
	\item We propose a novel end-to-end trainable architecture that can infer 3D hand poses from binary-valued silhouettes. Our method, which is called {\it Silhouette-Net}, requires minimal information to achieve comparable results as most of the RGB-based and depth-based approaches.
	
	\item We incorporate the notion of depth perception into our architecture as a latent representation, which provides sufficient details for hand pose estimation. The depth perception guidance is essential in our network since it automatically learns the required constraints and the spatial structure information.
	
	\item This work introduces a new task of estimating 3D hand poses from silhouettes. Our binary-based approach is competitive with the state-of-the-art depth-based methods, and should be able to serve as a strong baseline on this task for future study.
\end{enumerate}

% Related work needs a little pruning 
\section{Related Work}

\begin{figure*}[t]
	\centering
	\includegraphics[width=1\textwidth]{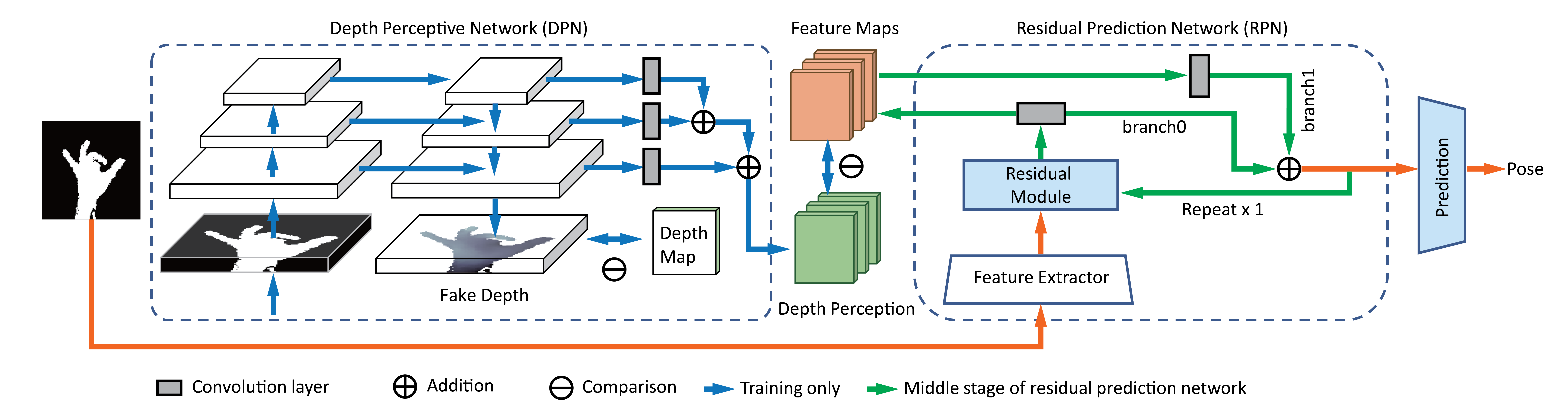}
	\caption{Our pipeline takes hand silhouettes as the input and predicts the 3D hand articulations. Depth Perceptive Network (DPN) generates a depth perception guidance $\Phi_\mathrm{dp}$ which is a multi-scale feature map that represents viewpoints and spatial distributions of hand joints. Residual Prediction Network (RPN) leverages the depth perception and the ground-truth hand pose to make accurate prediction. The overall network is an end-to-end structure with jointly trainable DPN and RPN. Note that the depth-related paths (in blue) are only active during training.} 
    \label{model}
\end{figure*}

\paragraph{3D hand pose estimation from RGB images:}
Early attempts~\cite{de2011model} have shown to be able to extract hand poses from RGB image sequences. More recent methods aim to estimate the 3D hand pose from a single RGB image using deep neural networks~\cite{zimmermann2017learning,tome2017lifting} owing to the advance in deep learning techniques and computation power. Some approaches apply the detection-based strategy~\cite{cai2018weakly} to a single RGB image, and then lift the predicted 2D heatmaps into 3D directly. Since the structure information obtained from a single RGB image is limited, Sridhar et al. \cite{sridhar2013interactive} adopt multi-view RGB images and depth data to estimate the 3D hand pose in a discriminative manner. Spurr et al.~\cite{spurr2018cross} propose a fully generative VAE framework to obtain a latent space, which is directly used to estimate hand poses from RGB images. 

\paragraph{3D hand pose estimation from depth maps:} 
The performance of estimating 3D hand poses from depth maps has been improved rapidly~\cite{choi2015collaborative,deng2017hand3d,ye2016spatial,baek2018augmented,wan2018dense} in terms of prediction accuracy. The studies on depth-based hand pose estimation generally adopt either generative or discriminative methods. Some generative methods track a hand pose through a manually designed hand model and a matching function by fitting the depth data~\cite{sridhar2016real,tkach2016sphere}, while the discriminative methods directly learn a hand pose from the annotated 3D data. Before the recent approaches exploit deep networks and the availability of considerable labeled data~\cite{yuan20172017}, some early methods focus on mapping between depth data and hand poses~\cite{keskin2012hand}. More recently, the state-of-the-art approaches further address several aspects such as the dimension of data in 2D or 3D~\cite{baek2018augmented,ge20173d}, the detection or regression based methods, etc. Most of the depth-based approaches adopt a single-view depth map, while some methods generate 3D voxels~\cite{moon2018v2v} or depend on the multiple viewpoints of orthogonal planes~\cite{ge2016robust} to use the 3D spatial information for pose estimation. Some detection-based methods use Recombinator Networks~\cite{honari2016recombinator} to produce a probability dense map for each hand joint. Regression-based methods ~\cite{madadi2017end,chen2017pose,ge2018hand} transform the locations of hand joints into the depth map, e.g., Region Ensemble Network~\cite{guo2017towards} and DeepPrior++~\cite{oberweger2017deepprior++}, which improve the accuracy by data augmentation and better initial hand localization. 

\paragraph{3D hand pose estimation from silhouettes:}
Not much previous research has focused on the estimation of 3D pose from silhouettes. Dibra et al. \cite{dibra2017human} develop a method for capturing human body shape from a human silhouette. However, to the best of our knowledge, no promising work has been presented on solving 3D hand pose estimation from silhouettes. Therefore, inspired by the aforementioned progress in 3D hand pose estimation using deep learning, we develop a novel network structure for learning to infer 3D hand poses from silhouettes, basing on the advantage of end-to-end training to achieve similar or better performance than the more common depth-based and RGB-based approaches.

%===========================================================================

\hfill
\section{Proposed Framework}  

% Seems we cannot ~ref section in AAAI, comment this temporarily 
%We first give an overview of our method in Section~\ref{sec:overview}. The network architecture is introduced in Section~\ref{sec:architecture}. Finally, Section~\ref{sec:training_inference} details the procedures of training and inference.

%\subsection{Overview}
%\label{sec:overview}
% our goal, whole network overview
The proposed network architecture, {\em Silhouette-Net}, is an end-to-end trainable network whose objective is to predict the 3D hand pose based on binary-valued silhouettes only. A 3D hand pose is represented by a set of hand articulations $\mathcal{X} = \{ \mathbf{x}_j \in \mathbb{R}^3\}_{j=1}^{J}$, where $J$ is the number of hand joints ($J=21$ in this work) and $\mathbf{x}_j$ denotes the 3D position of joint $j$. Figure~\ref{model} illustrates the proposed network architecture. Silhouette-Net consists of two sub-networks: Depth Perceptive Network (DPN) and Residual Prediction Network (RPN), which attain different objectives. Given the cropped image of hand silhouette posing a certain gesture, the overall network aims to resolve the difficulty of lacking the depth cue and to deal with the underlying learning task. The output of the network comprises the 3D locations of hand joints, which are represented in the real-world coordinates and scale (measured in millimeter) and are evaluated under the standard error metrics suggested by the benchmark dataset HIM2017~\cite{yuan20172017}. Note that our DPN is only used during training. At the testing phase, we only need RPN to predict hand joints without using the depth cue.  

% DPN, imitates depth-based methods but with an additional stage concerning the generative model that converts binary images to depth images.  
We deploy and modify the Feature Pyramid Network (FPN) framework \cite{LinDGHHB17} to discover the latent depth space through the learning process with depth guidance. The modified network, Depth Perceptive Network, is abbreviated to DPN as a sub-network in our Silhouette-Net architecture. DPN is trained under supervision with depth maps, and is crucial to facilitate the training of Silhouette-Net since it generates reliable guidance for disambiguating the variability in hand shape with respect to the silhouette. In addition, our experiments show that DPN can improve the performance in 3D hand pose estimation as it provides additional constraints on the latent hand pose space.

% RPN
Residual Prediction Network (RPN) leverages the advantage of detection-based method and regression-based method by inducing the probability feature map and performing direct regression. RPN branches into two main paths and eventually predicts the hand joint locations: {\em branch0} extracts the higher level features and {\em branch1} produces a latent heatmap based on the feedback from the guidance. In our design, DPN complements the preceding RPN properly during the training phase within the end-to-end trainable architecture.

%===========================================================================

\subsection{Network Architecture of Silhouette-Net}
\label{sec:architecture}

\paragraph{Feature Pyramid Network (FPN):} Features of multiple spatial resolutions can be extracted using FPN, which contains a top-down pathway with lateral skip-connections. The top-down pathway upsamples the received feature map and integrates it with higher-resolution features from the bottom-up pathway. We modify the FPN structure to learn the \textit{depth perception}, which is a latent representation that downscales to $1/2$ or $1/4$ of the original resolution of input image.

\paragraph{Depth Perceptive Network (DPN):} In our design, we seek to learn the depth perception guidance as a latent representation that contains multi-scale knowledge of depth by an FPN-style mechanism. For the purpose of achieving accurate prediction from silhouettes, the learned guidance representation should {\em i}) extract features that contain effective spatial information to localize the hand articulations, {\em ii}) combine multi-resolution features to learn the concept of depth (i.e., depth perception), and {\em iii}) be able to approximate the depth map to ensure that there is sufficient influence from the depth perception guidance. We design a lightweight FPN-style module which consists of three levels of downsampling and upsampling. The highest level of pyramid produces a feature at the $1/8$ scale, and the other two levels operate at the scale of $1/4$ and $1/2$. We employ a similar mechanism as in FPN: for each upsamping stage, adding a higher-level downsampling skip-connection in the output of the previous stage, where each upsampling stage performs $3\times 3$ deconvolution, batch normalization, ReLU, and a $3\times 3$ convolution with skip-connection. 

The objective of DPN is to provide the guidance for RPN and to address the problem of inferring spatial information from hand silhouettes. The depth perception guidance $\Phi_\mathrm{dp}$ covers multi-scale resolution and combines the output of each level of upsampling. We adopt a unique transformation for each scale of information to form the same channel dimension, where the transformation is mainly implemented by convolution layers with respect to the specified resolution. The depth perception guidance $\Phi_\mathrm{dp}$ is therefore a tensor of size $S \times S \times (J \cdot V)$, where $(J\cdot V)$ is the number of channels accounting for the $J$ hand joints multiplied by $V$ views. In our experiment $S$ is fixed to the scale of $1/4$ of the hand silhouette and $V$ is 1 or 3 since we consider single-view or three-view input silhouettes. Given the hand silhouettes of $V$ views, DPN produces a depth perception $\Phi_\mathrm{dp}$ with $(J \cdot V)$ channels.

We model a depth generator path (decoder) that begins from the lowest level to the highest level of upsampling to generate a fake depth map $\mathbf{D}_\mathrm{fake}$. The generated depth map is supervised by the real depth map $\mathbf{D}_\mathrm{depth}$, which is then represented as a guidance for the residual prediction network. The depth generator path aims to ensure that each scale of information contains sufficient and effective spatial cues. This additional constraint is optional and contains variants in our experiments \ref{sec_compare}, and can be written as the following loss function $\mathcal{L}_p$:
\begin{equation}
    \label{loss_P}
    \mathcal{L}_P = \| \mathbf{D}_\mathrm{fake} - \mathbf{D}_\mathrm{depth} \|_1 \,.
\end{equation}

\begin{figure}[t]
	\centering
	\includegraphics[width=0.45\textwidth]{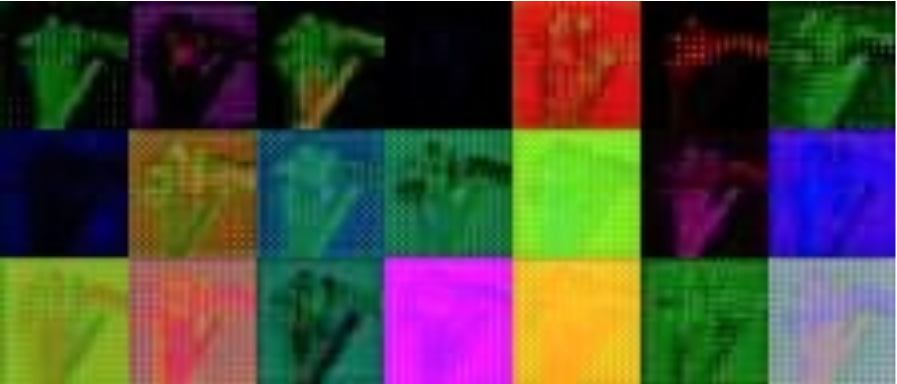}
	\caption{Visualization of the latent representation yielded by DPN. The resolution of depth perception is fixed to the scale of $1/4$ of an input image. The latent representation that models the depth perception can be seen as a useful guidance for RPN. This figure shows the 63 channels of depth perception as 21 color images.} 
    \label{results_depth_perception}
\end{figure}

DPN provides RPN a latent heatmap for estimating the locations of hand joints. The depth perception heatmap is a combination of features at each level of the pyramid, and optionally including the generated fake depth map $\mathbf{D}_\mathrm{fake}$. In our experiments, we show that the combination of multi-level resolutions performs better than explicitly using the generated fake depth map as guidance. A visualization of depth perception is shown in Figure~\ref{results_depth_perception}. The number of channels is set to be the number of joints times the number of views.

\paragraph{Residual Prediction Network (RPN):} The objective of RPN is to infer the 3D hand joint from the binary images of hand silhouettes. Inspired by Wu et al. \cite{Wu18HandPose}, we employ an effective residual architecture that leverages the advantages of detection-based and regression-based methods. We first apply a feature extractor that contains three convolution layers and a residual module similar to Inception ResNet~\cite{szegedy2017inception}. In the middle stage of RPN, it receives higher-level features generated by the residual module and comprises two branches. One branch (branch 0) keeps the extracted higher-level features with the information of hands, and the other branch (branch 1) is guided by the depth perception $\Phi_\mathrm{dp}$ to automatically constrain the predicted heatmap in the proper feature domain. At the final stage of RPN, two branches are merged and element-wise summed to yield the input for the prediction block.
The prediction block directly regresses our predicted 3D coordinates $\hat{\mathbf{x}}_{j}$ to the ground truth $\mathbf{x}_{j}$ by minimizing the L2 loss $\mathcal{L}_\mathrm{reg}$:
\begin{equation}
    \label{loss_regre}
    \mathcal{L}_\mathrm{reg}(\mathbf{x}_{j}, \hat{\mathbf{x}}_{j}) =  \sum_{j=1}^{J}\| \mathbf{x}_{j} - \hat{\mathbf{x}}_{j} \|_{2}^{2} \,.
\end{equation}

The middle stage of RPN is designed to predict a proper latent heatmap $\mathbf{H}$ with $(J \cdot V)$ channels, which can be supervised by depth perception $\Phi_\mathrm{dp}$. Each channel of the heatmap can be viewed as a pixel-wise probability for each hand joint. Compared with the human designated constraints or the guidance map, the network can learn the accurate prediction by automatically inferring the depth perception. We apply the smooth L1 loss $\mathcal{L}_\mathbf{dp}$ for the supervision between depth perception guidance $\Phi_{dp}$ and the produced latent heatmap $\mathbf{H}$ as follows:
\begin{equation}
\label{loss_dp}
\mathcal{L}_\mathrm{dp}(\mathbf{H}, \Phi_\mathrm{dp}) = \widetilde{\ell_1}\left(\|\mathbf{H}- \Phi_\mathrm{dp}\|_1 \right) \,,
\end{equation}
in which 
\begin{equation}
\widetilde{\ell_1}(z) =
    \begin{cases} 
     z-0.5\,, & |z|>1\,, \\
    0.5 z^2\,, & \mbox{ otherwise}. 
    \end{cases}
\end{equation}
We also adopt the identity in the middle stage of RPN (repeated one time) to increase the confidence of heatmap $\mathbf{H}$ as suggested in \cite{newell2016stacked}.

\begin{figure*}[t]
	\centering
	\includegraphics[width=0.95\textwidth]{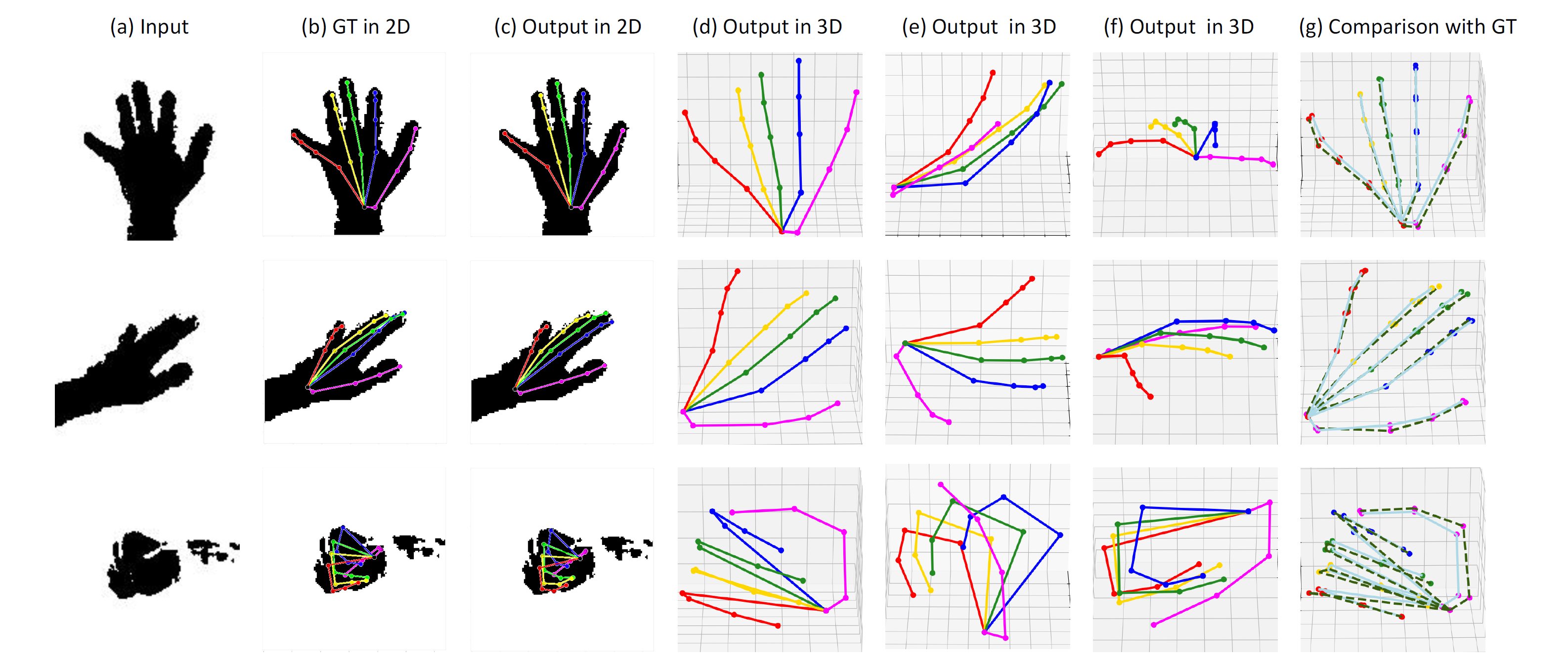}
	\caption{Experimental settings and results. (a) Hand silhouettes as the input. (b) Ground-truth 3D annotation projected to 2D. (c) Our predictions projected to 2D. (d) Our predictions in 3D. (e) Side views of our predictions in 3D. (f) Top views of our predictions in 3D. (g) Differences between the ground-truth joints (solid blue lines) and our predictions (dotted green lines).} 
    \label{hand_model}
\end{figure*}

\subsection{Training and Inference}
\label{sec:training_inference}

\paragraph{Training of Silhouette-Net:} By combining the losses in equations (\ref{loss_P}), (\ref{loss_regre}), and (\ref{loss_dp}), we obtain the overall loss function as $\mathcal{L}$:
\begin{equation}
    \mathcal{L} = \mathcal{L}_\mathrm{reg} + \lambda_{P}\mathcal{L}_{P} + \lambda_\mathrm{dp}\mathcal{L}_\mathrm{dp} + \lambda_W\mathcal{L}_W \,,
    \label{eq:loss}
\end{equation}
where $\mathcal{L}_\mathrm{dp}$ from (\ref{loss_dp}) is for the depth perception guidance that supervises the latent heatmap generated at the middle stage of RPN, and $\mathcal{L}_{P}$ from (\ref{loss_P}) ensures that the learned depth perception should contain the sufficient spatial information concerning the depth. We apply a common $\mathcal{L}_W$ regularization term to the weights of convolution operators. For these loss terms, $\lambda_{P}$, $\lambda_\mathrm{dp}$, and $\lambda_W$ are the scaling factors.

\paragraph{Inference of Silhouette-Net:} We predict the 3D hand pose from merely the hand silhouettes in our inference pipeline (illustrated in Figure~\ref{model}), where only the residual prediction module takes part in the prediction task without the supervision of either the depth map or the depth perception guidance. The following loss function $\mathcal{L}_I$ indicates the inference process:
\begin{equation}
    \mathcal{L}_I = \mathcal{L}_\mathrm{reg} \,.
\end{equation}
The inference of Silhouette-Net shows the possibility of using minimal information to infer 3D human hand articulations. With the mechanism of depth perception guidance in the proposed architecture during training, our model is able to achieve better or similar performance on the evaluation benchmark to the state-of-the-art depth-based approaches.

\section{Experiments}

\subsection{Experimental Setup}

The performance is evaluated using the \textit{2017 Hands In the Million Challenge} (HIM2017)~\cite{yuan20172017} dataset, which is created by sampling images from BigHand2.2M~\cite{yuan2017bighand2} and First-Person Hand Action~\cite{garcia2018first}. HIM2017 contains approximately 950K raw depth images at $640\times 480$-pixel resolution. The dataset is fully annotated with 21 hand joints as the ground truth. We split the data into 8:1:1 for training/validation/test sets, respectively, and evaluate the 3D localization accuracy on all 21 annotated hand joints. The evaluation protocol we take is derived from and consistent with the benchmark used in Wu et al. ~\cite{Wu18HandPose}. We use this protocol in our comparison with state-of-the-art depth-based approaches.

The primary objective of our experiments is to predict the 3D hand pose solely from minimal information, i.e., hand silhouettes, see Figure~\ref{hand_model}. Therefore, we apply preprocessing as follows: align with the center of all the joints, crop the hand segments from the raw depth images, and then convert them into binary hand silhouettes without further data augmentation. Note that, in this work, we do not concern hand segmentation and detection, but mainly focus on demonstrating the effectiveness of depth perception guidance in non-depth-based hand pose estimation.

The experiments consider two input types: {\em i}) a 2D binary image of a single-view silhouette, which is resized to $128\times 128$ pixels, and {\em ii}) binary images of multi-view silhouettes (also resized to $128\times 128$ pixels) to provide the additional side-view information.
In the case of multi-view hand silhouettes, we reproject the 3D points (derived from depth images) onto three orthogonal planes as implemented in ~\cite{ge2016robust} and then convert the projected depth images into binary images.

The proposed pipeline, either takes single-view or multi-view silhouettes as input, adopts the guidance from depth perception $\Phi_\mathrm{dp}$ during training. Besides, we choose different resolution settings for the depth perception guidance of the single-view and multi-view approaches. We have the resolution in scale of $1/4$ ($32\times 32$ pixels) for single-view and scale of $1/2$ ($64\times 64$ pixels) for multi-view.

\subsection{Implementation Details}

Adam optimization with default parameters is employed for training. For the training phase, the whole network iterates $30$ epochs with a starting learning rate of $10^{-2}$ and the weight decay of $9\times 10^{-1}$. The scaling factors for Eq. (\ref{eq:loss}) are  $\lambda_{P}=0.1$, $\lambda_\mathrm{dp} = 0.1$, and $\lambda_W = 0.01$. Our method is implemented in TensorFlow, and all the experiments are performed on a GeForce GTX 1080 Ti.

\begin{figure}[t]
	\centering
	\includegraphics[width=0.495\textwidth]{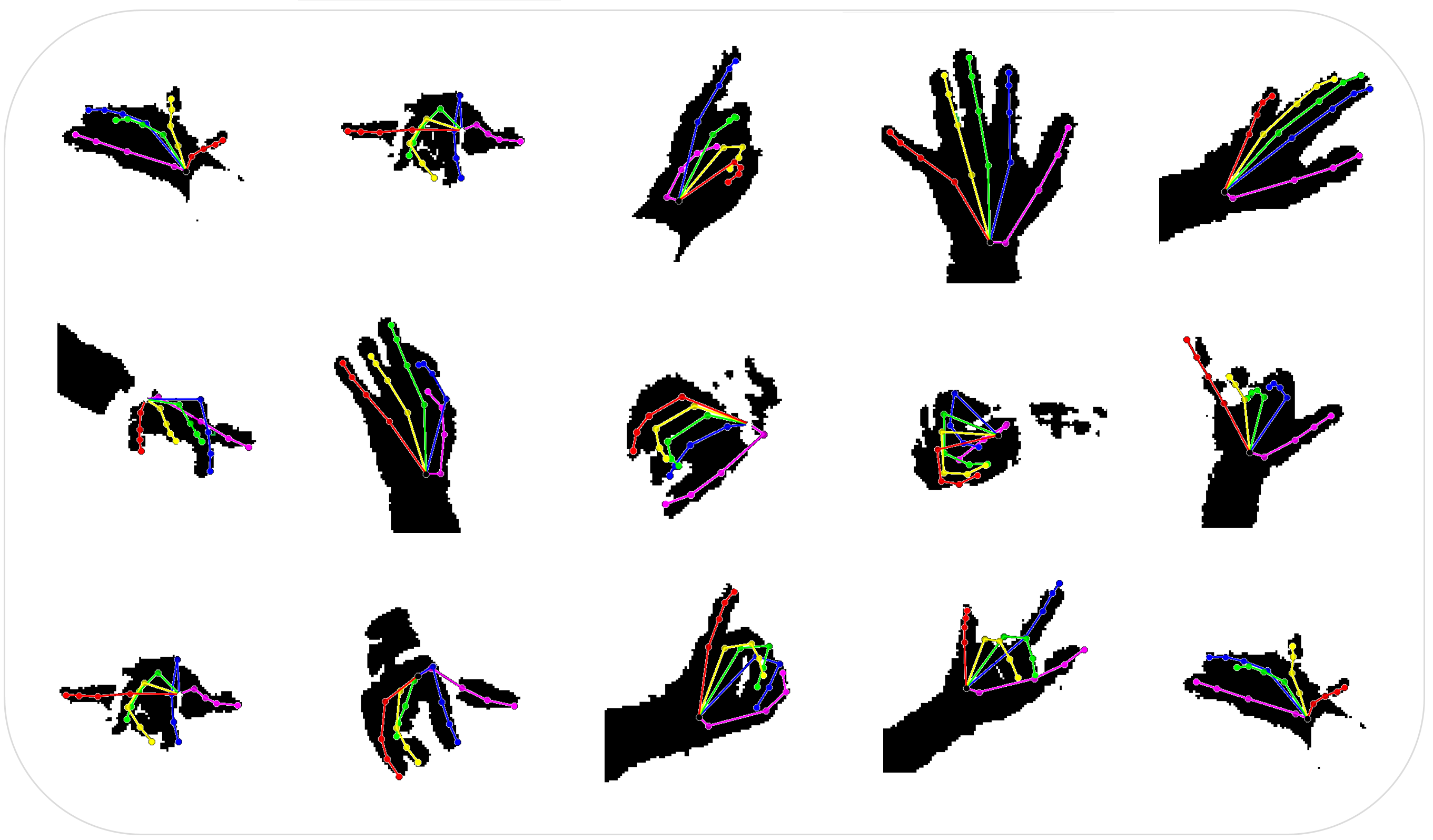}
	\caption{Some more examples of 3D pose estimation results. The predictions are projected to 2D view on top of the input silhouettes.}
    \label{results_3}
\end{figure}

\subsection{Evaluation Metrics}

We report the standard error metrics for all the joints in all testing frames of \textit{2017 Hands In the Million Challenge} ~\cite{yuan20172017} in terms of {\em mean error} and {\em max-per-joint error} in real-world coordinates (measured in millimeter). Since there is few related research which can be found concerning 3D hand pose estimation from binary hand data, our evaluation results are mainly compared with depth-based approaches notwithstanding that we use less input information. To show the validity of the comparison results, the test set for evaluating depth-based approaches and our Silhouette-Net architecture is consistent.

\subsection{Silhouette-Net and Its Variants}
\label{sec_compare}
In this section, we show our evaluation results of 3D hand pose estimation from hand silhouettes using Silhouette-Net and its variants. Figures~\ref{hand_model}~\&~\ref{results_3} illustrate some qualitative results obtained by Silhouette-Net.

\begin{table}[b] % Ablation
  \centering
  \caption{The {\em mean error} and {\em max-per-joint error} on different implementations of depth perception $\Phi_\mathrm{dp}$, where the number in parentheses denotes the max-per-joint error. DP: implementation type of depth perception $\Phi_\mathrm{dp}$, half-level (HDP) or full-level (FDP). Fake Depth: guidance including the generated depth map $D_\mathrm{fake}$.}\smallskip
      \begin{tabular}{|l|c|c|l|}
      \hline
      Model & DP & Fake Depth &  Reported Errors \\
      \hline\hline
      Baseline & \xmark & \cmark & 7.10 (10.32) \\
      Baseline & HDP & \cmark & 6.17 (9.18) \\
      Baseline & FDP & \cmark  & 5.81 (7.90) \\
      Silhouette-Net & FDP & \xmark & \textbf{5.60 (7.63)} \\
      \hline
      \end{tabular}
\label{ablation}
\end{table}

% Compare with state-of-the-art-method

\begin{figure}[t]
	\centering
	\includegraphics[width=\linewidth]{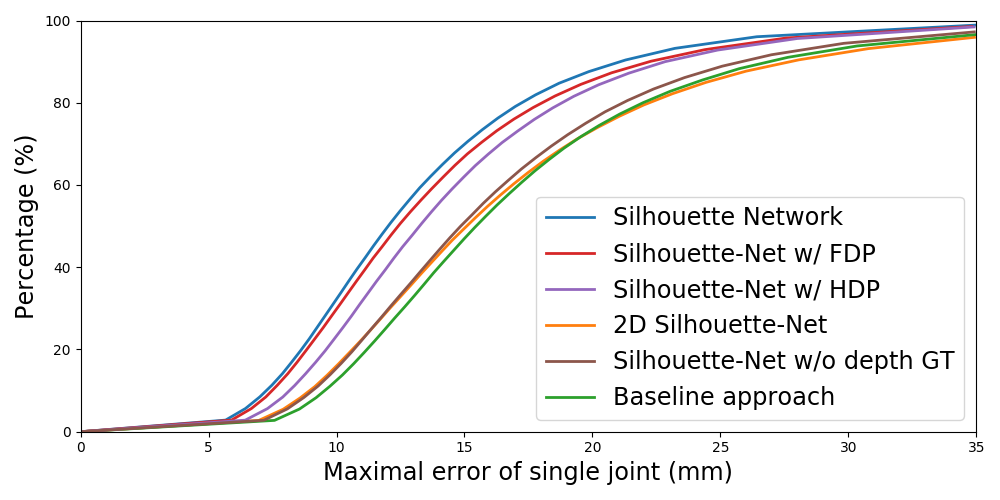}
	\caption{A comparison of our method and the variants. We illustrate the effectiveness of depth perception in different combinations.} 
	\label{err_rate}
\end{figure}

\begin{figure}[t]
    \centering
	\includegraphics[width=\linewidth]{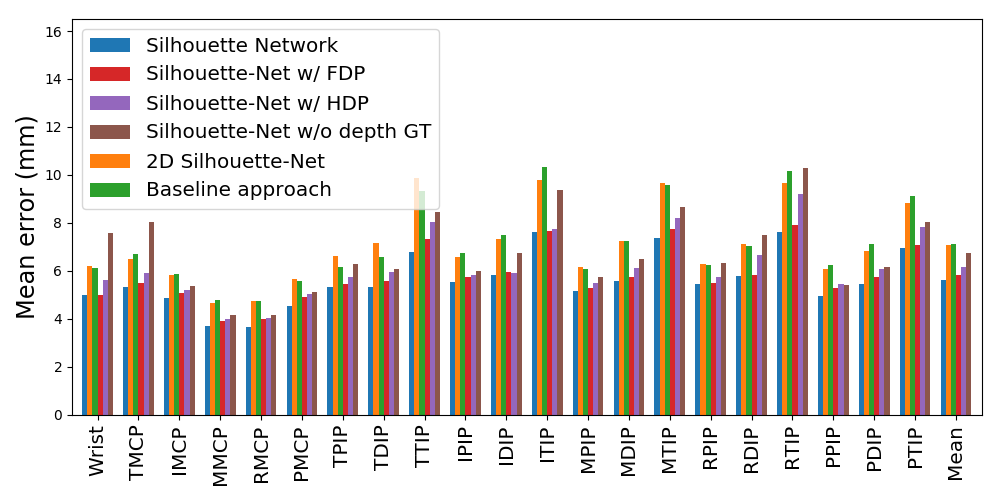}
	\caption{The mean error comparison between predicted hand joints, where T, I, M, R, and P denote the thumb, index, middle, ring, and pinky fingers. Silhouette-Net with depth perception guidance performs best when using full level of multi-resolution information without the generated depth map.} 
    \label{err_bar}
\end{figure}

\paragraph{Learning depth perception guidance:} To demonstrate the validity of learning depth perception guidance, we analyze the performance of our baseline (intuitive method) and several variants of depth perception as illustrated in Table~\ref{ablation}. For fair comparison, we adopt the identical RPN with the three-view setting for the experiment. We design an intuitive yet effective approach as our baseline, which comprises a generative model (encoder-decoder framework) and an RPN without the guidance of $\Phi_\mathrm{dp}$. The DPN in the baseline is equivalent to a depth generator which generates a fake depth map $\mathbf{D}_{fake}$ and feeds it to the regression module.

As shown in Figure~\ref{err_rate} and Figure~\ref{err_bar}, we analyze the variants of combinations of depth perception, including half-level of depth perception (HDP) and full-level of depth perception (FDP), where ``level'' indicates the involved level of upsampling in DPN (from the highest level at scale of $1/8$ to the lowest level at scale of $1/2$). For instance, HPD constitutes the resolution at $1/8$ to $1/4$ and FPD consists of the resolution at $1/8$ to $1/2$ with respect to the input size. The studies suggest the effectiveness of the proposed DPN in modeling the latent space of depth perception and providing sufficient information. Furthermore, from the experimental results we can find that the proposed DPN performs best when the generated depth map $\mathbf{D}_{fake}$ is excluded. Such results imply that the actual depth values in the depth map might not be requisite for 3D hand pose estimation, while the implicitly learned spatial cues from the silhouettes are rather useful and sufficient.

% same as our argue in intro:
As mentioned earlier in the introduction and now further supported by the experimental results, the learning of depth perception guidance $\Phi_\mathrm{dp}$ is essential for receiving spatial information of hand pose estimation while the exact depth map might not be. The automatically learned depth perception guidance also demonstrates its generalizability, in comparison with the manually designed constraints~\cite{Wu18HandPose}. The version of depth perception with full-level of FPN is called Silhouette-Net, which achieves the best performance in terms of mean error ($5.60$ mm) and max-per-joint error ($7.63$ mm) on HIM2017 dataset.

\begin{table}[t]
 \centering
 \caption{A summary of 3D hand joint estimation error, evaluated on HIM2017 dataset. The variants of our method all use the hand silhouettes as the input rather than the depth map. Silhouette-Net: three-view input silhouettes using the proposed architecture with the guidance of depth perception; 2D Silhouette-Net: singe-view version. All errors are reported in millimeter. The results are sorted with respect to the mean error. }
 \smallskip
     \begin{tabular}{|l|c|c|}
      \hline
      Model & Mean & Max-per-joint   \\
      & Error & Error \\
      \hline\hline
      \textbf{Silhouette-Net} & \textbf{5.60} & \textbf{7.63} \\
      Wu et al.(ECCV'18)%~\cite{Wu18HandPose} 
      & 5.9 & 8.04 \\
      \textbf{2D Silhouette-Net} &\textbf{7.07} & \textbf{9.85} \\ 
      Ge et al.(CVPR'17)%~\cite{ge20173d} 
      & 8.0 & 12.30 \\ 
      Ge et al.(CVPR'16)%~\cite{ge2016robust} 
      & 8.15 & 12.17 \\ 
      Single-view-baseline  &8.93 & 12.88 \\
      Moon et al.(CVPR'18)%~\cite{moon2018v2v} 
      & 9.49 & 13.60 \\
      \hline
      \end{tabular}
\label{sota}
\end{table}

\begin{table}
  \caption{Influence of the use of ground-truth depth map GT during training. Depth supervision in DPN is optional; however, Silhouette-Net performs better with the supervision from ground-truth depth during training. Note that depth maps are not needed by all the variants during testing. Depth perception (DP) here is implemented with full level of depth perception. All errors are reported in millimeter.}
  \label{self-compare}
  \begin{center}
      \begin{tabular}{|l|c|c|c|}
      \hline
       DP & GT & Mean  & Max-per-joint \\
        & Supervision & Error & Error \\
      \hline\hline
       \xmark & \cmark & 7.10 & 10.32 \\
       \cmark & \xmark & 6.75 & 10.29 \\
       \cmark & \cmark & \textbf{5.60} & \textbf{7.63} \\
      \hline                        
      \end{tabular}
  \end{center}
\end{table}

\paragraph{Multi-view vs. single-view:} To obtain more accurate and robust estimation, we preprocess the hand silhouettes to three views~\cite{ge2016robust}, in which the frontal view (single view) and the side-views complement each other. For the single-view approach, we also implement a simple network which is similar to Inception-Resnet ~\cite{szegedy2017inception} as the baseline. This single-view baseline employs 2D direct regression to predict 3D coordinates of joints without involving depth perception guidance and depth maps. Our 2D Silhouette-Net is implemented for single-view hand silhouette, which infers from a single binary hand image that contains only the frontal view. 
Table~\ref{sota} summarizes the evaluation results and comparisons of our Silhouette-Net and 2D Silhouette-Net versus the baseline and several state-of-the-art depth-based approaches.
As the results show, our single-view 2D Silhouette-Net is quite effective, yet the multi-view version Silhouette-Net can further improve the accuracy by a large margin, illustrating a promising solution to the problem of 3D hand pose estimation.

\paragraph{Supervision of ground-truth depth map:} In Silhouette-Net, the ground truth of depth map is optional during training (See Table \ref{self-compare}). The experiments suggest that the depth map is not as crucial as in depth-based method for our architecture; nevertheless, the performance can be further improved under the depth map supervision during training.

\subsection{Comparison with State-of-the-Art Depth-Based Methods}

We compare our best results with several state-of-the-art depth-based methods~\cite{Wu18HandPose,ge20173d,moon2018v2v} on HIM2017 benchmark. It can be seen that on HIM2017, even we use the minimal information of input data (binary-valued silhouettes), our end-to-end approach that automatically learns the depth perception is able to outperform the state-of-the-art depth-based method~\cite{Wu18HandPose} as shown in Table \ref{sota}. Our method improves the errors of (mean error, max-per-joint error) from (5.90, 8.04) to (5.60, 7.63). Note that our approach demonstrates the potential capability of using minimal information rather than relying on dense depth maps. Silhouette-Net is also superior to some of previous RGBD-based methods. 
%The improvement may be regarded as the depth perception guidance models the ambiguity that a hand silhouettes must cause and the hidden constraint of inter-joints of hand as well.

%===========================================================================
\section{Conclusion}

We have presented a new method to predict the 3D hand pose from hand silhouettes. The proposed architecture, {\em Silhouette-Net}, combines the Residual Prediction Network (RPN) with the generative Depth Perceptive Network (DPN), and we jointly train the entire model end-to-end using 2D binary hand images. The DPN learns from the latent depth space that contains sufficient information to guide and constrain the prediction of RPN. Silhouette-Net achieves comparable results of hand pose estimation as state-of-the-art depth-based methods by inferring the 3D hand pose from merely binary-valued hand silhouettes. 

\bibliographystyle{aaai} \bibliography{main}

\end{document}